\documentclass[onecolumn]{main}

\paperstyle{fancy}
\papercolor{black}
\input{frame-06}

\usepackage{amsmath,amsfonts,bm}

\def\eqref#1{equation~\ref{#1}}

\def\algref#1{algorithm~\ref{#1}}

\def\1{\bm{1}}

\DeclareMathAlphabet{\mathsfit}{\encodingdefault}{\sfdefault}{m}{sl}
\SetMathAlphabet{\mathsfit}{bold}{\encodingdefault}{\sfdefault}{bx}{n}

\hypersetup{hidelinks}

\def\ljm{\textcolor{black}}
\def\zyf{\textcolor{black}}
\def\hry{\textcolor{black}}

\title{FreshVLA: \textbf{F}requency-\textbf{R}egularized Tim\textbf{e} \textbf{S}cale \textbf{H}ierarchical Latent Actions
  for Vision-Language-Action Models}

\author[1,2,\ast]{Jiaming Liang}
\author[1,\ast]{Yifeng Zhan}
\author[1,\ast]{Rongyi He}
\author[2,3,\ast]{Mu Li}
\author[1]{Jiaqi Fan}
\author[1]{Zhiyuan Zhang}
\author[1]{Qiwei Liang}
\author[1]{Zhengjie Zhang}
\author[1]{Boyang Cai}
\author[2,\dagger]{Ningyuan Zhang}
\author[4]{Xingjian Jing}
\author[1,\ddagger]{Qiang Nie}

\affiliation[1]{The Hong Kong University of Science and Technology (Guangzhou)}
\affiliation[2]{XPENG Robotics}
\affiliation[3]{Shanghai Institute of Intelligent Science and Technology, Tongji University}
\affiliation[4]{City University of Hong Kong}

\contribution[\ast]{Equal Contribution}
\contribution[\dagger]{Project Lead}
\contribution[\ddagger]{Corresponding Author}

\abstract{
Learning reusable latent actions from action-free videos offers an effective route to improving Vision-Language-Action (VLA) policies. However, reconstruction-oriented latent actions do not necessarily benefit policy learning: as they may encode task-irrelevant static appearance with scene identity, or entangle slowly varying task progress with fine-grained execution dynamics in limited discrete code. We propose \textbf{FreshVLA}, a VLA framework built on a frequency-guided, dynamics-sensitive learning paradigm for acquiring temporally hierarchical continuous latent actions. FreshVLA uses a high-pass prior to guide visual transition encoding toward dynamical regions, thereby reducing reliance on static appearance cues. It further organizes latent actions into coarse-grained task progress and fine-grained execution dynamics through cross-timescale sampling and frequency-domain regularization, enhancing their capacity to represent long-horizon behaviors. During policy training, we freeze the pretrained latent action encoder and transfer its representations to robot control through joint flow-matching of latent and robot actions. FreshVLA achieves 98.2\% average success on LIBERO and 98.0\% on LIBERO-Long, exceeding GR00T-N1.6 on the latter by 3.6\%. It also reaches an average sequence length of 4.52 on CALVIN and strong overall success rate of 80.4\% in the zero-shot LIBERO-PLUS evaluation. Furthermore, FreshVLA exhibits greater robustness to visual perturbations and a more distinct division of roles between two branches in modeling low- and high-frequency motion, respectively. In 35-step action regression, it achieves a 51.3-percentage-point higher relative error reduction than its hierarchical discrete counterpart. Real-world experiments demonstrate that FreshVLA improves the average success rate over $\pi_{0.5}$ by 7.5\%. The FreshVLA project page is available at \url{https://freshvla.github.io/}.
}

\keywords{latent actions, vision-language-action models, temporal frequency}

\begin{document}
\maketitle

\section{Introduction}
Vision-Language-Action (VLA) models aim to translate language-conditioned visual understanding into executable robot behavior. Recent systems have advanced this paradigm through larger multimodal backbones, broader robot datasets, and more expressive action decoders~\citep{brohan2023rt2,ghosh2024octo,kim2024openvla,black2024pi0}. Nevertheless, a fundamental interface challenge remains: policies must transform visual intent into temporally coherent control behavior. This requires representing the visual transitions that should occur in a scene, including which object should move, how contact relationships evolve, and how local execution supports longer-horizon task progress.

Latent Action Models (LAMs) provide a natural means of learning such representations. By predicting or reconstructing observation transitions, LAMs infer latent actions from adjacent frames or observation sequences~\citep{ye2025lapa,liang2025clam,bu2025univla}. These variables can provide supervision for policy pretraining while reducing the need to directly align raw action spaces across embodiments. However, reconstruction objectives do not guarantee that latent actions capture controllable and task-relevant changes. They may instead encode background layouts, static appearance, or other cues that help predict future observations but are only weakly related to control~\citep{bu2025laof,cai2025spaceMotion,bu2026whatMatters}. \ljm{Furthermore, manipulation trajectories contain both slowly evolving task progress, such as approaching and transporting an object, and local execution dynamics, such as grasp adjustments and release timing. Many prior LAMs represent fixed-interval visual transitions with vector-quantized codes~\citep{vandenOord2017vqvae,ye2025lapa,bu2025univla}, or continuous formulations with a single latent transition scale~\citep{liang2025clam,yang2025como}, which may obscure this multi-timescale structure and hinder downstream policies from modeling the temporal dependencies of long-horizon manipulation.}

\begin{figure}[t]
\centering
\includegraphics[width=\linewidth]{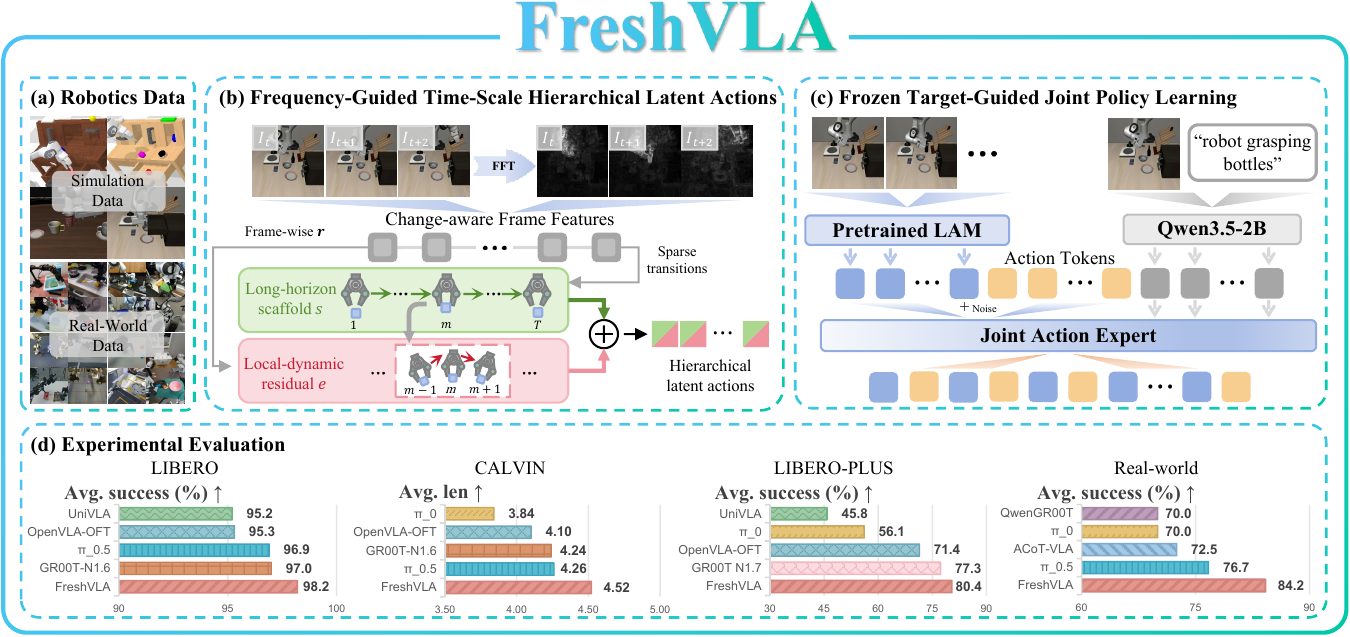}\\
\setlength{\abovecaptionskip}{2pt}
\caption{\ljm{FreshVLA uses a high-pass prior to extract dynamics-sensitive features and organizes them into hierarchical latent actions that capture long-horizon task progress. For policy learning, a frozen latent action encoder guides the joint denoising of latent and robot actions. Evaluations in simulation, representation analyses, and real-world manipulation demonstrate its effectiveness.}}
\label{fig:freshvla_overview}
\end{figure}

 \ljm{To address the aforementioned challenges, we propose \textbf{FreshVLA} to construct dynamics-sensitive hierarchical latent actions from visual transitions, as is described in Figure~\ref{fig:freshvla_overview}. In the visual feature space, temporal high-pass responses provide localized cues of change and guide the patch-level attention module toward regions exhibiting task-relevant visual dynamics. In the latent action space, temporal downsampling and upsampling, together with temporal-frequency regularization, encourage the substage-scale transition to model persistent long-horizon dependencies, while the execution-scale representation captures immediate dynamic residuals. Both representations are further used in bidirectional future prediction to explain changes in visual observations. During VLA learning, the pretrained latent action model is kept frozen to preserve the transition space learned from action-free videos. Conditioned on visual-language context, a latent-action flow-matching mechanism jointly models latent actions and robot actions}~\citep{lipman2023flowMatching,li2026rotvla}.

\ljm{With 2B parameters, FreshVLA achieves an average success rate of 98.2\% on LIBERO, reaches 98.0\% on LIBERO-Long, and attains an average sequence length of 4.52 on CALVIN. These results exceed those of GR00T-N1.6~\citep{nvidia2025grootn16} by 1.2 and 3.6 percentage points on the two LIBERO metrics, respectively, and by 0.28 subtasks on CALVIN, highlighting its strength in long-horizon manipulation. In zero-shot LIBERO-PLUS evaluation, it achieves 80.4\% macro-average success and outperforms OpenVLA-OFT~\citep{kim2025openvlaOft} across all seven perturbation categories, reaching 88.9\%, 97.0\%, and 97.1\% under language, lighting, and background shifts, respectively. Representation analyses further show that FreshVLA’s latent actions are more robust to visual perturbations and exhibit complementary frequency preferences across temporal scales. On the 35-step action-regression probe, FreshVLA achieves a relative error reduction 51.3-percentage-points higher than that of its hierarchical discrete counterpart. Across four real-world tasks, FreshVLA improves mean success by 7.5 percentage points over $\pi_{0.5}$, suggesting more reliable object interaction and long-horizon execution.}

\textbf{Our contributions are summarized as follows:}
\begin{itemize}
    \item We introduce a temporally hierarchical continuous latent action representation that organizes visual transitions into substage-scale components capturing long-horizon task progress and execution-scale residuals preserving local dynamics.

    \item We propose \textbf{FreshVLA}, which learns this representation using a temporal high-pass visual prior and frequency-domain regularization. A frozen latent action encoder then provides targets for VLA policy training through joint flow matching over latent and robot actions.

    \item FreshVLA achieves strong results on LIBERO and CALVIN and generalizes well to perturbations in zero-shot LIBERO-PLUS evaluations. Additional experiments and real-world deployments further demonstrate its effectiveness on long-horizon manipulation tasks.
\end{itemize}

\setlength{\textfloatsep}{8pt plus 2pt minus 2pt}

\section{Related Work}

\subsection{Vision--Language--Action Models}\hry{Vision--Language--Action (VLA) models map multimodal context---including visual observations, language instructions, and available proprioceptive states---to executable robot control. RT-1 and RT-2 established scalable language-conditioned robot-policy learning~\citep{brohan2023rt1,brohan2023rt2}. Subsequent generalist policies scale this paradigm by integrating heterogeneous robot data or pretrained multimodal backbones with specialized action generators~\citep{oneill2024openx,ghosh2024octo,kim2024openvla,black2024pi0,nvidia2025grootn1}. Despite their improved generality, these policies are trained to predict robot actions whose coordinates and control conventions remain embodiment dependent~\citep{oneill2024openx,bu2025univla}. At the level of labeled action representation, FAST applies DCT-based compression for efficient tokenization of high-frequency action chunks~\citep{pertsch2025fast}, whereas MINT decomposes action chunks in frequency space into coarse intent and fine execution~\citep{huang2026mint}. \ljm{While these methods improve the modeling of labeled robot actions, they remain embodiment-specific and require action supervision. This motivates learning embodiment-agnostic, control-oriented representations from action-free videos.}}

\subsection{Latent Action Learning}\hry{Action-free videos provide object-motion and task-progress cues at scale without robot action annotations~\citep{ye2025lapa,yang2025como}. Latent Action Models (LAMs) convert such visual changes into action-like variables, typically by inferring compact transition representations from observation pairs and using them for prediction or downstream control. Discrete approaches commonly employ VQ-VAE-style quantization~\citep{vandenOord2017vqvae,bruce2024genie}. LAPA applies such codes to VLA pretraining~\citep{ye2025lapa}, while UniVLA derives task-centric latent actions for cross-embodiment policy learning~\citep{bu2025univla}. Other work explores continuous latent transitions, representation alignment, and transfer from heterogeneous human videos~\citep{liang2025clam,yang2025como,liu2026lara,dai2026conla,zhang2026clap,luo2026jala}. Related formulations introduce image-goal, motion-token, distilled, unified world-model, or 3D-aware representations~\citep{chen2024igor,chen2025moto,li2025latbot,bi2026motus,govind2026unilact,Liang_2026_CVPR}, while a 2026 study evaluates 41 latent action design choices spanning modeling, objectives, and downstream integration~\citep{bu2026whatMatters}.} \ljm{FreshVLA leverages a high-pass prior to guide transition encoding toward dynamically changing regions, while cross-timescale construction and DCT-based regularization organize latent actions into substage-scale task progress and execution-scale local dynamics before embodiment-specific policy alignment.}

\section{Method}

\subsection{Overview and Problem Setup}

 A basic latent action model consists of an inverse dynamics model $\mathcal{E}$ and a forward dynamics model $\mathcal{D}$~\citep{ye2025lapa,liang2025clam}. Given consecutive observations $o_t$ and $o_{t+1}$, the inverse dynamics model extracts a latent transition $\tilde{z}_{t\rightarrow t+1}$, while the forward dynamics model predicts the next observation conditioned on the current observation and the inferred transition:

\begin{equation}
\tilde z_{t\rightarrow t+1}
=\mathcal E(o_t,o_{t+1}),
\qquad
\hat o_{t+1}
=\mathcal D(o_t,\tilde z_{t\rightarrow t+1}),
\qquad
\mathcal L_{\mathrm{LAM}}
=\left\|\hat o_{t+1}-o_{t+1}\right\|_2^2.
\end{equation}

\ljm{Because pronounced ego-motion from wrist-mounted cameras can be easily confounded with genuine interaction dynamics, FreshVLA currently pretrains its latent actions on fixed third-person videos, providing a stable observation coordinate frame. We further apply video stabilization and filter out frames with abrupt background or illumination changes to reduce nuisance variations that could interfere with the high-pass motion prior.}

\subsection{Stage I: Frequency-Guided Hierarchical Latent Targets}

\begin{figure}[t]
\centering
\includegraphics[width=\linewidth]{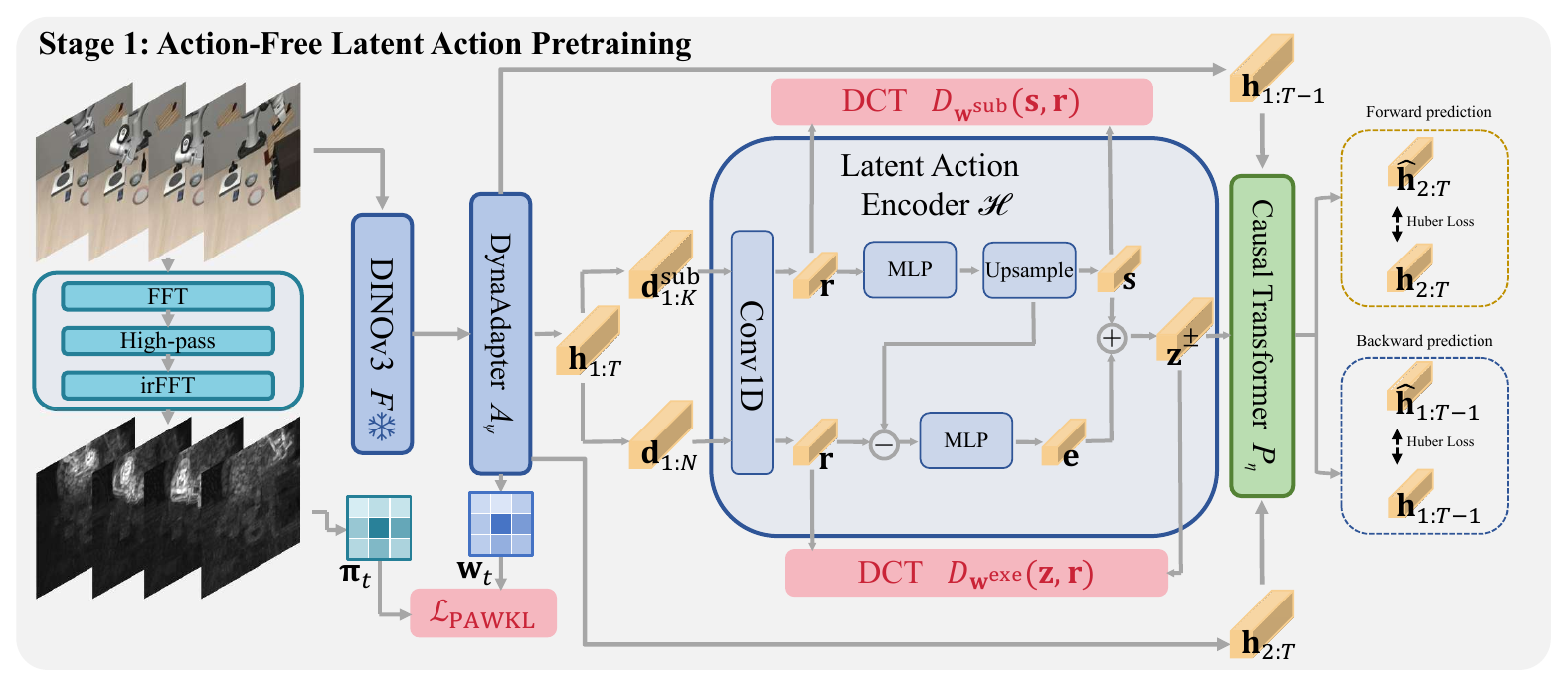}\\
\setlength{\abovecaptionskip}{2pt}
\caption{Illustration of the pretraining architecture. DINOv3 and DynaAdapter first extract dynamics-sensitive visual transitions under FFT-based PAWKL guidance. The latent action encoder $\mathcal{H}$ then constructs temporally hierarchical continuous latent actions, while DCT regularization encourages scale-specific temporal structure, with Causal Transformer as a bidirectional predictor.}
\label{fig:hcla}
\end{figure}

\textbf{Dynamics-Sensitive Visual Transition Encoding.} As discussed above, LAMs may place excessive attention on static information such as background appearance, resulting in predictive shortcuts and misleading action representations. We therefore introduce a high-pass response prior to guide the model toward dynamically changing regions involved in manipulation. As illustrated in Figure~\ref{fig:hcla}, given an observation window $\mathbf{o}_{1:T}$, a frozen DINOv3 visual encoder~\citep{simeoni2025dinov3} $F$ extracts patch-level tokens $\mathbf{U}_t$. A trainable DynaAdapter $A_\psi$ aggregates these tokens into frame-level representations $\mathbf{h}_t$ and outputs normalized patch-attention distributions $\mathbf{w}_t$. We then compute frame-level transition features by differencing adjacent frame-level representations:

\begin{equation}
\begin{aligned}
\mathbf{U}_t &= F(\mathbf{o}_t),\quad
(\mathbf{h}_t,\mathbf{w}_t)=A_\psi(\mathbf{U}_t),\quad
\mathbf{d}_t = \mathbf{h}_{t+1}-\mathbf{h}_t.
\end{aligned}
\end{equation}

\ljm{Specifically, we apply a real-valued FFT~\citep{cooley1965fft} along the temporal axis at each RGB pixel and retain frequency bins $k\geq\lfloor0.5(\lfloor T/2\rfloor+1)\rfloor$. The inverse FFT yields frame-wise high-pass responses, whose magnitudes are normalized and aggregated across RGB channels by taking the maximum. After spatial pooling onto the patch grid, the resulting scores are normalized to form target distributions $\boldsymbol{\pi}_t$, with gradients stopped through the target branch. For a batch of $B$ observation windows indexed by $b$, the Patch-Attention Weight KL Alignment loss is defined as}

\begin{equation}
 \mathcal{L}_{\mathrm{PAWKL}}
=
\frac{1}{BT}
\sum_{b=1}^{B}\sum_{t=1}^{T}
D_{\mathrm{KL}}
\left(
\operatorname{sg}(\boldsymbol{\pi}_{b,t})
\,\middle\|\,
\mathbf{w}_{b,t}
\right).
\end{equation}

\ljm{PAWKL encourages the learned attention distribution to approach the temporal high-pass target, thereby reducing the visual aggregator’s reliance on static appearance and directing attention toward regions undergoing visual change.} 

\textbf{Hierarchical Latent Action Construction.} Many existing latent action models rely on flat VQ codebooks~\citep{ye2025lapa,bu2025univla}, \ljm{which limit their ability to express long-horizon relationships, so we introduce a temporally hierarchical continuous representation. Let $N=T-1$ denote the number of frame-level transitions in an observation window of $T$ frames.} Approximately equally spaced temporal anchors $\alpha_1<\cdots<\alpha_{K+1}$ divide each observation window into $K$ coarse intervals. The substage-scale input summarizes the accumulated feature change within each interval:

\begin{equation}
\mathbf{d}^{\mathrm{sub}}_k
=
\mathbf{h}_{\alpha_{k+1}}
-
\mathbf{h}_{\alpha_k},
\qquad
k=1,\ldots,K. 
\end{equation}

We use a shared temporal encoder to project both transition types into a common hidden space and model temporal dependencies at different scales. The substage transition encoder $R_s$ captures long-horizon task dependencies and is then upsampled to construct a residual signal. The execution transition encoder $R_e$ processes this residual to obtain the execution-scale representation:

\begin{equation}
\label{eq:hierarchical_latent}
\begin{aligned}
\mathbf{r} &= E_c(\mathbf{d}_{1:N}),\qquad
\mathbf{s} = \operatorname{Up}_{N}
\left(R_s\left(E_c(\mathbf{d}^{\mathrm{sub}}_{1:K})\right)\right),\\
\mathbf{e} &= R_e(\mathbf{r}-\mathbf{s}),\qquad
\mathbf{z} = \operatorname{LN}_{\mathrm{out}}(\mathbf{s}+\mathbf{e}).
\end{aligned} 
\end{equation}

\ljm{Overall, Eq.~\ref{eq:hierarchical_latent} defines the latent action encoder} $\mathcal{H}$:
$\mathbf{z}=\mathcal{H}(\mathbf{d}_{1:N},\mathbf{d}^{\mathrm{sub}}_{1:K})$. Here, \ljm{$\mathbf{r},\mathbf{s},\mathbf{e},\mathbf{z}\in\mathbb{R}^{B\times N\times D}$, where $D$ is the feature dimension.} Specifically, $\mathbf{r}$ denotes the encoded frame-level transition \ljm{relationship}, $\mathbf{s}$ the upsampled substage-scale representation, $\mathbf{e}$ the execution-scale representation, and $\mathbf{z}$ the composite hierarchical latent action.

\ljm{To explicitly encourage the temporal-scale structure in Eq.~\ref{eq:hierarchical_latent}, we align $\mathbf{s}$ and $\mathbf{z}$ with  $\mathbf{r}$ in the DCT domain~\citep{ahmed1974dct}. The substage alignment uses the first $K$ temporal frequency bins, encouraging $\mathbf{s}$ to capture slowly varying components of $\mathbf{r}$. We align $\mathbf{z}$ over all $N$ bins to promote fidelity across temporal frequencies, with scale-specific frequency weights defined below. For two such sequences $\mathbf{x},\mathbf{y}\in\mathbb{R}^{B\times N\times D}$, let $\mathcal{C}(\cdot)$ denote the orthogonal DCT applied along the temporal axis independently for each feature channel. We define the weighted discrepancy as}

\begin{equation}
D_{\boldsymbol{\omega}}(\mathbf{x},\mathbf{y})
=
\frac{1}{BD}
\sum_{b=1}^{B}
\sum_{\ell=1}^{D}
\sum_{j=0}^{N-1}
\omega_j
\left|
\mathcal{C}(\mathbf{x})_{bj\ell}
-
\mathcal{C}(\mathbf{y})_{bj\ell}
\right|^2.  
\end{equation}

\ljm{The temporal DCT maps each length-$N$ transition sequence to $N$ frequency bins indexed by $j=0,\ldots,N-1$. The substage weights $\boldsymbol{\omega}^{\mathrm{sub}}$ assign normalized exponentially decaying weights to the first $K$ bins, whereas the execution weights $\boldsymbol{\omega}^{\mathrm{exe}}$ apply the same decay rule over all $N$ bins. To learn transitions in both temporal directions, we apply the shared hierarchical encoder in Eq.~\ref{eq:hierarchical_latent} to the original and sign-reversed transition features. The resulting forward and reverse representations are denoted by $(\mathbf{r}^{\sigma},\mathbf{s}^{\sigma},\mathbf{e}^{\sigma},\mathbf{z}^{\sigma})$, where $\sigma\in\{+,-\}$. The hierarchical encoding objective is}

\begin{equation}
\mathcal{L}_{\mathcal{H}}
=
\frac{1}{2}
\sum_{\sigma\in\{+,-\}}
\left[
\lambda_{\mathrm{sub}}
D_{\boldsymbol{\omega}^{\mathrm{sub}}}
\left(
\mathbf{s}^{\sigma},\mathbf{r}^{\sigma}
\right)
+
\lambda_{\mathrm{exe}}
D_{\boldsymbol{\omega}^{\mathrm{exe}}}
\left(
\mathbf{z}^{\sigma},\mathbf{r}^{\sigma}
\right)
\right].    
\end{equation}

\textbf{Bidirectional Prediction Objective.}
\ljm{The forward and reverse latent actions are given by}

\begin{equation}
\mathbf{z}^{+}
=
\mathcal{H}
\left(
\mathbf{d},
\mathbf{d}^{\mathrm{sub}}
\right),
\qquad
\mathbf{z}^{-}
=
\mathcal{H}
\left(
-\mathbf{d},
-\mathbf{d}^{\mathrm{sub}}
\right).  
\end{equation}

A shared causal Transformer $P_\eta$ predicts adjacent frame representations conditioned on hierarchical latent actions. \hry{For reverse prediction, $\operatorname{rev}(\cdot)$ temporally reverses inputs, latents, and targets. The training losses are}

\begin{equation}
\begin{gathered}
\mathcal{L}_{\mathrm{dyn}}
=
\frac{1}{2}\Big[
\ell_{\mathrm{Huber}}
\left(
P_\eta(\mathbf{h}_{1:T-1},\mathbf{z}^{+}),
\mathbf{h}_{2:T}
\right)
+
\ell_{\mathrm{Huber}}
\left(
P_\eta(
\operatorname{rev}(\mathbf{h}_{2:T}),
\operatorname{rev}(\mathbf{z}^{-})
),
\operatorname{rev}(\mathbf{h}_{1:T-1})
\right)
\Big],
\\
\mathcal{L}_{\mathrm{total}}
=
\lambda_{\mathrm{d}}\mathcal{L}_{\mathrm{dyn}}
+
\lambda_{\mathrm{h}}\mathcal{L}_{\mathcal{H}}
+
\lambda_{\mathrm{p}}\mathcal{L}_{\mathrm{PAWKL}}
+
\lambda_{\mathrm{s}}\mathcal{L}_{\mathrm{SIG}}.
\end{gathered}
\end{equation}

\ljm{The prediction objective uses the Huber loss~\citep{huber1964robust}. Here, $\mathcal{L}_{\mathrm{SIG}}$ is the Sketched Isotropic Gaussian regularizer introduced in LeJEPA~\citep{balestriero2025lejepa}, applied to the frame-level representations $\mathbf{h}_{1:T}$.} It encourages the empirical characteristic function of randomly projected features to approach that of a standard Gaussian distribution, thereby mitigating representation collapse.

\setlength{\textfloatsep}{8pt plus 2pt minus 2pt}

\subsection{Stage II: Joint Latent-Action Flow Matching}

Building on flow-matching~\citep{lipman2023flowMatching}, we apply the hierarchical latent actions to VLA policies through a \ljm{joint latent action and robot action training paradigm.} After Stage I, we freeze the visual encoder $F$, DynaAdapter $A_\psi$, and latent action encoder $\mathcal{H}$ as a target generator $G$. Given an observation window, $G$ produces forward hierarchical latent actions:

\begin{equation}
G(\mathbf{o}_{t:t+T-1})
=
\mathcal{H}
\left(
\mathbf{d}_{1:N},
\mathbf{d}^{\mathrm{sub}}_{1:K}
\right)
=
\mathbf{z}^{+}.
\end{equation}

Let $H_{\mathrm{a}}\leq N$ denote the action horizon. The first $H_{\mathrm{a}}$ latent actions, $\mathbf{z}^{+}_{1:H_{\mathrm{a}}}$, are aligned with the corresponding robot-action chunk $\mathbf{a}_{t:t+H_{\mathrm{a}}-1}$. The policy is conditioned on the current observation, language instruction, \ljm{and robot state}, denoted by $\mathbf{c}_t=(\mathbf{o}_t,l,\mathbf{q}_t)$. Both streams are independently perturbed with Gaussian noise and jointly processed by a shared flow-matching action expert:

\begin{equation}
\widetilde{\mathbf{z}}_{\tau,i}
=
(1-\tau)\boldsymbol{\epsilon}^{z}_{i}
+
\tau\mathbf{z}^{+}_{i},
\qquad
\widetilde{\mathbf{a}}_{\tau,i}
=
(1-\tau)\boldsymbol{\epsilon}^{a}_{i}
+
\tau\mathbf{a}_{t+i-1},
\qquad
i=1,\ldots,H_{\mathrm{a}}.
\end{equation}

\ljm{Here, $\boldsymbol{\epsilon}^{z}_{i}$ and $\boldsymbol{\epsilon}^{a}_{i}$ refer to standard Gaussian noise. The action expert predicts velocity fields $\widehat{\mathbf{v}}^{a}_{\tau,i}$ and $\widehat{\mathbf{v}}^{z}_{\tau,i}$ for the robot-action and latent-action streams, respectively, with the objective}

\begin{equation}
\mathcal{L}_{\mathrm{JFM}}
=
\mathbb{E}
\left[
\frac{1}{H_{\mathrm{a}}}
\sum_{i=1}^{H_{\mathrm{a}}}
\left(
\operatorname{MSE}
\left(
\widehat{\mathbf{v}}^{a}_{\tau,i},
\mathbf{a}_{t+i-1}
-
\boldsymbol{\epsilon}^{a}_{i}
\right)
+
\operatorname{MSE}
\left(
\widehat{\mathbf{v}}^{z}_{\tau,i},
\mathbf{z}^{+}_{i}
-
\boldsymbol{\epsilon}^{z}_{i}
\right)
\right)
\right].
\end{equation}

Joint denoising transfers visual-dynamics information from the latent action stream to continuous robot action prediction. During inference, $G$ and future observations are unnecessary. Both streams are initialized with Gaussian noise and jointly denoised, while only the resulting robot action chunk is executed; the latent stream remains an internal variable.

\section{Experiments}

\subsection{Experimental Setup}

\textbf{Dataset and Training Configuration.}
We construct a large-scale pretraining corpus of 227,894 trajectories from diverse real and simulated environments. For latent action pretraining, hierarchical continuous latent actions are extracted from 36-frame observation windows. For downstream policy training, we use Qwen3.5-2B~\citep{qwen2026qwen35} as the vision-language backbone and a DiT-based joint flow-matching action expert to generate 7-D robot actions. Please refer to Appendix~\ref{subsec:app_architecture} for architecture details and Appendices~\ref{subsec:app_dataset} and ~\ref{subsec:app_training_details} for data and training configurations.

\subsection{Simulation Experiments}

\textbf{LIBERO.} We evaluate FreshVLA on LIBERO~\citep{liu2023libero}, a widely used language-conditioned robotic manipulation suite for assessing policy generalization under diverse task variations. Following the standard protocol, we report results on four suites: LIBERO-Spatial, LIBERO-Object, LIBERO-Goal, and LIBERO-Long. We compare against LAPA~\citep{ye2025lapa}, OpenVLA~\citep{kim2024openvla}, OpenVLA-OFT~\citep{kim2025openvlaOft}, VILLA-X~\citep{chen2026villax}, UniVLA~\citep{bu2025univla}, $\pi_0$~\citep{black2024pi0}, $\pi_{0.5}$~\citep{physicalintelligence2025pi05}, JALA~\citep{luo2026jala} and GR00T-N1.6~\citep{nvidia2025grootn16}. As shown in Table~\ref{tab:libero_and_calvin_results}, FreshVLA achieves an average success rate of 98.2\% with 2B parameters and outperforms these strong baselines overall. Although FreshVLA underperforms on LIBERO-Goal, likely because its latent action pretraining does not explicitly condition on language goals, it remains highly competitive overall. Notably, FreshVLA achieves a 98.0\% success rate on LIBERO-Long, improving over UniVLA and GR00T-N1.6 by 6.0 and 3.6, respectively. This result suggests that temporally hierarchical latent actions better preserve the compositional structure required for long-horizon manipulation.

\begin{table}[t]
\centering
\small\renewcommand{\arraystretch}{1.15}
\caption{Performance on LIBERO and CALVIN. LIBERO reports success rates
(\%) for each suite and their average. CALVIN reports success rates (\%)
for completing at least $k$ consecutive subtasks ($k=1,\ldots,5$) and
the mean number of completed subtasks (Len.). ``w/o LA'' denotes the
FreshVLA variant trained without latent action integration.}
\label{tab:libero_and_calvin_results}
\setlength{\tabcolsep}{2pt}
{%
\begin{tabular}{@{}lc*{5}{c}*{6}{c}@{}}
\toprule
\multirow{2}{*}{Method} & \multirow{2}{*}{Size}
  & \multicolumn{5}{c}{\textbf{LIBERO}} & \multicolumn{6}{c}{\textbf{CALVIN}} \\
\cmidrule(lr){3-7} \cmidrule(lr){8-13}
  & & Spatial & Object & Goal & Long & Avg.
    & 1 & 2 & 3 & 4 & 5 & Len. \\
\midrule
LAPA~\citep{ye2025lapa}
  & 7B & 73.8 & 74.6 & 58.8 & 55.4 & 65.7
  & -- & -- & -- & -- & -- & -- \\
OpenVLA~\citep{kim2024openvla}
  & 7B & 84.7 & 88.4 & 79.2 & 53.7 & 76.5
  & 91.3 & 77.8 & 62.0 & 52.1 & 43.5 & 3.27 \\
OpenVLA-OFT~\citep{kim2025openvlaOft}
  & 7B & 96.2 & 98.3 & 96.2 & 90.7 & 95.3
  & 96.3 & 89.1 & 82.4 & 75.8 & 66.5 & 4.10 \\
VILLA-X~\citep{chen2026villax}
  & 3B & 97.5 & 97.0 & 91.5 & 74.5 & 90.1
  & -- & -- & -- & -- & -- & -- \\
UniVLA~\citep{bu2025univla}
  & 9B & 96.5 & 96.8 & 95.6 & 92.0 & 95.2
  & 95.5 & 85.8 & 75.4 & 66.9 & 56.5 & 3.80 \\
$\pi_0$~\citep{black2024pi0}
  & 3B & 96.8 & 98.8 & 95.8 & 85.2 & 94.2
  & 93.8 & 85.0 & 76.7 & 68.6 & 60.1 & 3.84 \\
$\pi_{0.5}$~\citep{openpi2025libero}
  & 3B & \textbf{98.8} & 98.2 & \textbf{98.0} & 92.4 & 96.9
  & 94.4 & 88.4 & 85.3 & 80.1 & 76.1 & 4.26 \\
JALA~\citep{luo2026jala}
  & - & 96.0 & 98.2 & 97.4 & 96.0 & 96.9
  & -- & -- & -- & -- & -- & -- \\
GR00T-N1.6~\citep{nvidia2025grootn1}
  & 3B & 97.7 & 98.5 & 97.5 & 94.4 & 97.0
  & 96.3 & 91.4 & 85.3 & 79.7 & 71.7 & 4.24 \\
\midrule
\textbf{FreshVLA (Ours)}
  & 2B & 98.7 & \textbf{99.7} & 96.3 & \textbf{98.0} & \textbf{98.2}
  & \textbf{99.6} & \textbf{97.9} & \textbf{92.4}
  & \textbf{85.1} & \textbf{77.5} & \textbf{4.52} \\
\quad w/o LA
  & 2B & 98.7 & 96.0 & 96.3 & 93.0 & 96.0
  & 95.4 & 90.1 & 85.2 & 80.3 & 74.9 & 4.26 \\
\bottomrule
\end{tabular}%
}
\end{table}

\textbf{LIBERO-PLUS.} To further evaluate generalization under distribution shifts, we assess FreshVLA on LIBERO-PLUS~\citep{fei2026liberoPlus}. Built upon the standard LIBERO suites, LIBERO-PLUS is a robustness benchmark that introduces perturbations along seven dimensions. As shown in \zyf{Table~\ref{tab:libero-plus}}, FreshVLA achieves the best performance across six perturbation dimensions, with a macro-average success rate of 80.4\%, outperforming other baselines. In particular, FreshVLA shows clear advantages under language, light, background, and noise perturbations. These results suggest that the robustness of the learned latent actions to visual and semantic variations transfers to the downstream policy, thereby improving its out-of-distribution generalization.

\begin{table}[t]
\centering
\small
\caption{Zero-shot success rates (\%) on LIBERO-PLUS.
Avg.\ denotes the macro-average across the seven perturbation categories.
``w/o LA'' denotes the FreshVLA variant trained without latent action integration.}
\label{tab:libero-plus}
\setlength{\tabcolsep}{3pt}
\resizebox{0.8\linewidth}{!}{%
\begin{tabular}{@{}l*{8}{c}@{}}
\toprule
Method & Camera & Robot & Language & Light & Background & Noise & Layout & Avg. \\
\midrule
OpenVLA~\citep{kim2024openvla}
  & 0.8 & 3.5 & 23.0 & 8.1 & 34.8 & 15.2 & 28.5 & 16.3 \\
OpenVLA-OFT~\citep{kim2025openvlaOft}
  & 56.4 & 31.9 & 79.5 & 88.7 & 93.3 & 75.8 & 74.2 & 71.4 \\
UniVLA~\citep{bu2025univla}
  & 1.8 & 46.2 & 69.6 & 69.0 & 81.0 & 21.2 & 31.9 & 45.8 \\
$\pi_{0}$~\citep{black2024pi0}
  & 13.8 & 6.0 & 58.8 & 85.0 & 81.4 & 79.0 & 68.9 & 56.1 \\
GR00T N1.7 \cite{nvidia2025grootn1}
& 53.1 & \textbf{63.2} & 85.0 & 94.8 & 93.9 & 76.1 & 75.1 & 77.3 \\
\midrule
\textbf{FreshVLA (Ours)}
  & \textbf{60.0} & 58.9 & \textbf{88.9} & \textbf{97.0} & \textbf{97.1} & \textbf{84.4} & \textbf{76.3} & \textbf{80.4} \\
\quad w/o LA
  & 58.9 & 58.8 & 86.0 & 91.8 & 84.0 & 71.8 & 76.1 & 75.3 \\
\bottomrule
\end{tabular}%
}
\end{table}
\setlength{\textfloatsep}{10pt plus 2pt minus 2pt}

\textbf{CALVIN.} To evaluate a model’s ability to execute multiple robot manipulation tasks consecutively, we evaluate FreshVLA on CALVIN~\citep{mees2022calvin}. Following the standard protocol, each instruction sequence consists of five consecutive subtasks. As shown in Table~\ref{tab:libero_and_calvin_results}, FreshVLA achieves the best success rate at every sequence length, reaching 99.6\%, 97.9\%, 92.4\%, 85.1\%, and 77.5\% after completing 1--5 subtasks, while its first-subtask success rate exceeds $\pi_{0.5}$, by 5.2\%. FreshVLA also achieves the highest average completion length of 4.52. These results suggest that its temporally hierarchical latent actions help sustain reliable execution across extended task sequences.

\subsection{Representation Analysis Experiments}

\begin{figure}[t]
\centering
\includegraphics[width=\linewidth]{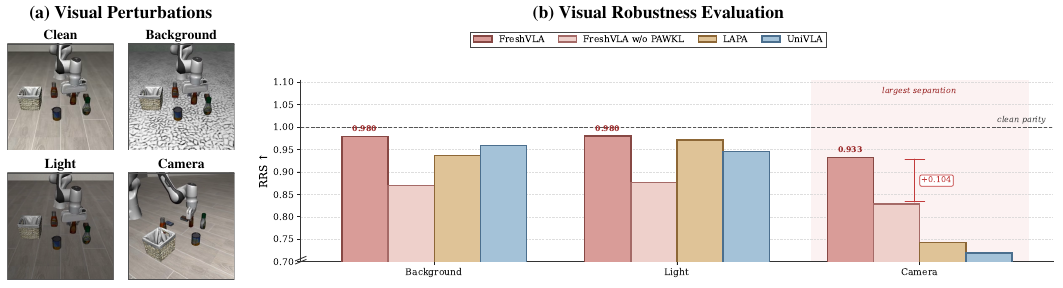}
\setlength{\abovecaptionskip}{2pt}
\caption{Visual robustness of latent actions.
\textbf{(a)} A clean observation and examples of background, lighting, and camera changes.
\textbf{(b)} Relative robustness scores for each perturbation, measured as the ratio of clean to perturbed action-prediction MSE.}
\label{fig:app_visual_perturbations}
\end{figure}

\textbf{Visual Robustness.} We study whether latent actions retain action-decodable information under visual shifts. Following LIBERO-PLUS, we perturb only the third-person-view images through background, illumination, or camera changes while keeping trajectories and action targets fixed (Figure~\ref{fig:app_visual_perturbations}(a)). We compare LAPA, UniVLA, FreshVLA, and FreshVLA without PAWKL. For each frozen model, the pretrained latent action encoder extracts latent actions from both clean and perturbed inputs, and the same MLP action regressor is trained on clean LIBERO trajectories and evaluated on perturbed inputs without adaptation, so that any degradation reflects perturbation sensitivity of the latent actions themselves. To isolate the model's intrinsic robustness to visual shifts from its base performance, we follow standard robustness evaluation protocols~\citep{imagenet-c} and report the Relative Robustness Score(RRS):

\begin{equation}
\mathrm{RRS}
=
\frac{
    \mathrm{MSE}_{\mathrm{clean}}
}{
    \mathrm{MSE}_{\mathrm{perturbed}}
}.
\label{eq:retention}
\end{equation}

As shown in Figure~\ref{fig:app_visual_perturbations}(b), FreshVLA achieves the highest RRS across all perturbations, reaching 0.980, 0.980, and 0.933 for Background, Light, and Camera, respectively. Camera changes cause the largest performance gap, indicating that viewpoint variations are particularly disruptive to latent action representations. Removing PAWKL consistently degrades it, \zyf{ which confirms that FreshVLA effectively preserves action-relevant information, decoupling it from visual distractors.}

\textbf{Long-Horizon Gains.}
To further analyze how latent action structure affects long-horizon task expression, we conduct an action-regression probe to measure the influence of different context on regression error. We freeze the latent action encoders of all four variants: SCLA/SDLA (single-layer continuous/discrete) and HCLA/HDLA (hierarchical continuous/discrete), then independently train an action regressor~\citep{yang2025como,actionprobe} on its latent representations. All regressors use the same transformer architecture, parameter count, data split, and optimization protocol, thereby controlling for differences in probe capacity. A local bidirectional mask limits the visible context to $H_{\mathrm{c}}$ steps. On held-out trajectories from the LIBERO benchmark, let $E(H_{\mathrm c})$ denote the held-out action-prediction MSE when the probe observes $H_{\mathrm c}$ latent steps. Taking the single-step error $E^{\mathrm{base}} := E(1)$ as the baseline, we define
\begin{equation}
\mathrm{RER}(H_{\mathrm c})
= \frac{E^{\mathrm{base}}-E(H_{\mathrm c})}
       {E^{\mathrm{base}}}.
\end{equation}

\begin{figure}[t]
    \centering
    \begin{minipage}[c]{0.48\textwidth}
        \centering
        \resizebox{\linewidth}{!}{%
        \begin{tabular}{lcccc}
        \toprule
        Suite & SDLA & HDLA & SCLA & \textbf{HCLA (Ours)} \\
        \midrule
        Long & 92.0 & 72.3 & 93.0 & \textbf{98.0} \\
        Goal & 96.0 & 96.0 & 95.7 & \textbf{96.3} \\
        Object & 96.7 & 93.3 & 99.7 & \textbf{99.7} \\
        Spatial & 95.3 & 90.7 & 95.7 & \textbf{98.7} \\
        \midrule
        Avg. & 95.0 & 88.1 & 96.0 & \textbf{98.2} \\
        \bottomrule
        \end{tabular}%
        }
    \end{minipage}
    \hfill
    \begin{minipage}[c]{0.48\textwidth}
        \centering
        \includegraphics[width=\linewidth]{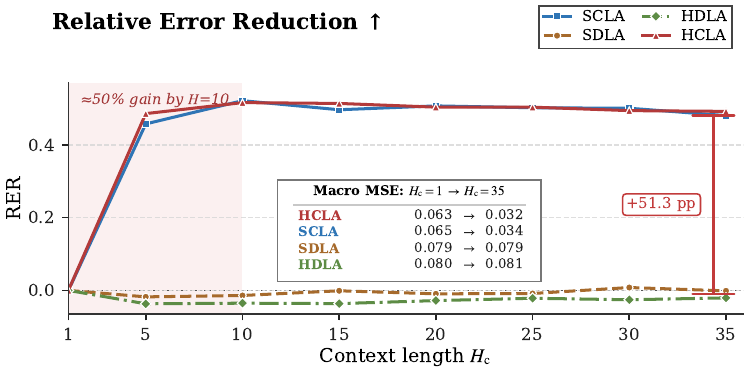}
\end{minipage}

\setlength{\abovecaptionskip}{2pt}
\caption{\textbf{Left:} Ablation study of latent action structures on LIBERO.
\textbf{Right:} Action-regression probe showing Relative Error Reduction (RER)
versus context length $H_{\mathrm{c}}$.}
\label{fig:ablation_rer_combined}
\end{figure}

\ljm{As shown in Figure~\ref{fig:ablation_rer_combined}(right), SDLA and HDLA improve only marginally as $H_{\mathrm{c}}$ increases, whereas SCLA and HCLA exhibit lower action-prediction error even at $H_{\mathrm{c}}=1$ and benefit markedly from additional context. At $H_{\mathrm{c}}=35$, HDLA incurs a substantially higher MSE than HCLA and achieves an RER that is 51.3\% lower, suggesting that continuous latent actions support more effective context-dependent action decoding. Despite similar RER trends, HCLA achieves approximately 5.9\% lower MSE than SCLA at $H_{\mathrm c}=35$, indicating an additional benefit from temporal hierarchy.}

\setlength{\textfloatsep}{10pt plus 2pt minus 2pt}

\textbf{Frequency Specialization.} During steady transport, successive end-effector pose increments are similar, so their temporal sequence concentrates energy at low frequencies. A rapid pose correction before grasping changes those increments over a few frames, introducing higher-frequency components. Inspired by this, we ask whether this distinction in latent actions is reflected in the robot motion that can be decoded from each branch. To test this prediction, we freeze the encoder and train four probes, one for each combination of branch and frequency band, to predict the corresponding DCT coefficients of 6-D end-effector pose increments. We compute predictive $R^2$ within each LIBERO suite using its mean coefficient vector as the baseline prediction, then average the scores across the four suites.

\begin{figure}[tbp]
\centering
\includegraphics[width=0.8\linewidth]{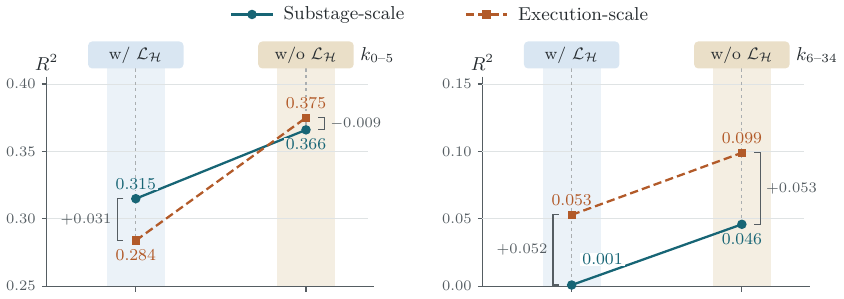}
\caption{{Predictive $R^2$ for end-effector motion, averaged across four LIBERO suites. The temporal DCT bin index $k$ ranges from 0 to 5 in the low band and from 6 to 34 in the high band. Teal circles and orange squares denote the substage and execution branches, respectively. Brackets show the $R^2$ difference: substage minus execution for low frequencies, and execution minus substage for high frequencies.}}
\label{fig:dct_loss_probe}
\end{figure}

Only with the DCT loss do we observe the expected frequency specialization (Figure~\ref{fig:dct_loss_probe}). The substage branch leads the execution branch by 0.031 in predictive $R^2$ in the low band, while the execution branch leads by 0.052 in the high band. Removing the loss raises all four probe scores, but the execution branch gains more in the low band and leads in both bands, by 0.009 and 0.053, respectively. This reversal in the low-frequency ordering is consistent with the direct low-frequency supervision applied to the substage representation. Despite better band-wise decoding, following protocol in Figure~\ref{fig:ablation_rer_combined} (right), we find that remove $\mathcal{L}_{\mathcal{H}}$ increased macro-averaged MSE from 0.032 to 0.033 and reduced RER from 49.22\% to 46.68\%. Together, these results suggest that $\mathcal{L}_{\mathcal{H}}$ encourages the two branches to specialize at different timescales and improves the accuracy of decoding long action sequences from the learned representations.

\subsection{Real-World Experiments}
To evaluate FreshVLA's transferability to real-world environments, we conduct experiments on a 6-DoF Flexiv robotic arm equipped with RealSense D435 cameras, as shown in Figure~\ref{fig:real_world_setup}. We evaluate four language-conditioned tabletop manipulation tasks spanning basic to long-horizon operations. For each task, we collect 30 trajectories and train four baselines under the same setting. The QwenGR00T baseline draws on Qwen3.5~\citep{qwen2026qwen35} and a GR00T-style action expert~\citep{nvidia2025grootn1}, while ACoT-VLA~\citep{zhong2026acotvla} and $\pi_{0.5}$~\citep{physicalintelligence2025pi05} are selected as strong reasoning- and long-horizon-oriented baselines, respectively. As reported in Table~\ref{tab:real_world_results}, FreshVLA achieves an average success rate of $84.2\%$, outperforming ACoT-VLA and $\pi_{0.5}$ by 11.7 and 7.5 percentage points, respectively. \ljm{On Warmup Dinner, which requires precise object interaction, FreshVLA presses the bell's raised button more reliably and achieves a higher success rate. We hypothesize that its latent actions' focus on interaction-relevant visual dynamics contributes to this improvement. FreshVLA also achieves success rates of 86.7\% and 73.3\% on the multistage, multiobject Organize Bowls and Collect Toys tasks, respectively. These results support the utility of temporally hierarchical continuous latent actions for multistage, contact-rich manipulation in real-world settings.}

\begin{table}[t]
\centering
\caption{Real-world manipulation success rates (\%) across four tasks.}
\label{tab:real_world_results}
\begin{tabular}{lccccc}
\toprule
Method & Open Pot & Warmup Dinner & Organize Bowls & Collect Toys & Avg. \\
\midrule
$\pi_0$          & 96.7 & 63.3 & 53.3 & 66.7 & 70.0 \\
QwenGR00T        & 93.3 & 60.0 & 66.7 & 60.0 & 70.0 \\
ACoT-VLA         & 96.7 & 70.0 & 56.7 & 66.7 & 72.5 \\
$\pi_{0.5}$       & 96.7 & 66.7 & 80.0 & 63.3 & 76.7 \\
\textbf{FreshVLA (Ours)}
                 & \textbf{96.7} & \textbf{80.0} & \textbf{86.7}
                 & \textbf{73.3} & \textbf{84.2} \\
\bottomrule
\end{tabular}
\end{table}

\subsection{Ablation Study}

\textbf{Latent Action Structure Analysis.} We compare four latent action configurations for downstream control. These include single-scale continuous (SCLA), single-scale discrete (SDLA), hierarchical continuous (HCLA), and hierarchical discrete (HDLA) representations. The hierarchical variants use the temporal downsampling, upsampling, and residual decomposition described in ~\eqref{eq:hierarchical_latent}, whereas the discrete variants use a VQ codebook to compress latent actions. As shown in Figure~\ref{fig:ablation_rer_combined} (left), HCLA achieves the best average success rate of 98.2\%. Within the continuous setting, temporal hierarchy improves LIBERO-Long performance by 5.0\% over the single-scale counterpart, supporting its utility for long-horizon manipulation.

\textbf{Effect of Temporal Scale.} The temporal granularity of hierarchical latent actions can significantly affect their representation quality. We therefore evaluate VLA performance with different numbers of substages under a fixed 36-frame input window. As the number of substages increases, the temporal span represented by the substage-scale latent action becomes shorter. As depicted in Table~\ref{tab:substage_number}, using 6 or 12 substages outperforms using 3, with 6 substages achieving the highest average success rate of 98.2\% and the best LIBERO-Long performance of 98.0\%. This suggests that a suitable temporal granularity improves the modeling of long-horizon task dynamics.

\begin{table}[t]
\centering
\caption{Ablation study of the number of substages on LIBERO.}
\label{tab:substage_number}
\begin{tabular}{lccccc}
\toprule
Substage Number & Long & Goal & Object & Spatial & Avg. \\
\midrule
3  & 90.0 & 97.3 & 97.3 & 93.7 & 94.6 \\
6  & \textbf{98.0} & 96.3 & 99.7 & 98.7 & \textbf{98.2}\\
12 & 94.3 & \textbf{98.3} & \textbf{100.0} & \textbf{99.3} & 98.0 \\
\bottomrule
\end{tabular}
\end{table}

\textbf{Frequency-Regularization Design.} We first isolate the effect of frequency regularization by removing $\mathcal{L}_{\mathcal{H}}$ while keeping the temporal hierarchy fixed. As shown in Table~\ref{tab:fr_decay_ablation}, removing the loss reduces mean LIBERO success from 98.2\% to 96.5\%, with the largest drop on LIBERO-Long, from 98.0\% to 93.0\%. We then vary the decay coefficient, which controls how quickly the loss weights decrease with frequency. Uniform weighting (decay 0) retains $\mathcal{L}_{\mathcal{H}}$ and achieves nearly the same mean success as the default exponential weighting (decay 1), at 98.1\% versus 98.2\%. Otherwise, uniform weighting performs better on LIBERO-Goal (98.3\% versus 96.3\%), while the default achieves higher success on LIBERO-Long (98.0\% versus 96.0\%). Increasing the coefficient to 2 lowers mean success to 97.7\%, despite achieving 100.0\% on LIBERO-Object. Overall, all three weighting schemes improve mean and long-horizon success over the unregularized model, with the default decay yielding the strongest long-horizon performance.

\textbf{Effectiveness of Latent Actions.} To understand the effect of latent actions to downstream control, we compare FreshVLA with a variant that predicts only robot actions. Table~\ref{tab:libero_and_calvin_results} demonstrate that jointly denoising latent and robot actions raises mean LIBERO success from 96.0\% to 98.2\%, with the largest gain on LIBERO-Long (93.0\% to 98.0\%). On CALVIN, the proportion of evaluation sequences in which the policy completes at least three consecutive subtasks rises from 85.2\% to 92.4\%, and the proportion in which it completes all five rises from 74.9\% to 77.5\% (Table~\ref{tab:libero_and_calvin_results}). These gains extend to all seven perturbation categories in LIBERO-PLUS, where macro-average success rises from 75.3\% to 80.4\%. The largest improvements occur under background changes, sensor noise, and lighting changes, with gains of 13.1, 12.6, and 5.2 percentage points, respectively (Table~\ref{tab:libero-plus}).

\textbf{Effectiveness of PAWKL Divergence.} To assess the contribution of temporal high-pass guidance, we compare FreshVLA variants trained with and without PAWKL on LIBERO and LIBERO-PLUS (Table~\ref{tab:pawkl_ablation}). On LIBERO, PAWKL raises mean success from 97.7\% to 98.2\%, with gains of 2.7 percentage points on LIBERO-Long . Success on LIBERO-Goal declines slightly, possibly because emphasizing interaction regions reduces sensitivity to goal-location cues. On LIBERO-PLUS, PAWKL raises macro-average success from 74.8\% to 80.4\%. Although the variant trained directly on DINOv3 features without PAWKL achieved a higher success rate under changes in object layout (82.4\% vs. 76.3\%), adding PAWKL improved success by 22.4 points under sensor noise, 13.3 points under camera viewpoint changes, and 8.8 points under changes to the robot’s initial state. Together, these results suggest that emphasizing manipulation-relevant visual changes benefits long-horizon control and helps the policy retain action-relevant information under several forms of perturbation.

\begin{table*}[!t]
  \centering
  \caption{Effect of PAWKL on LIBERO and zero-shot LIBERO-PLUS success
  rates (\%). Avg.\ denotes the mean across the four LIBERO suites or
  the seven LIBERO-PLUS perturbation categories. The higher success
  rate in each column is shown in bold.}
  \label{tab:pawkl_ablation}
  \small
  \setlength{\tabcolsep}{2pt}
  \begin{tabular*}{\textwidth}{
    @{\extracolsep{\fill}}l*{5}{r}@{\hspace{0.8em}}*{8}{r}@{}
  }
    \toprule
    & \multicolumn{5}{c}{\textbf{LIBERO}}
    & \multicolumn{8}{c}{\textbf{LIBERO-PLUS (zero-shot)}} \\
    \cmidrule(lr){2-6} \cmidrule(l){7-14}
    Variant
      & Spa. & Obj. & Goal & Long & Avg.
      & Cam. & Robot & Lang. & Light & Bg. & Noise & Layout & Avg. \\
    \midrule
    w/o PAWKL
      & 98.3 & 99.3 & \textbf{98.0} & 95.3 & 97.7
      & 46.7 & 50.1 & \textbf{90.5} & 95.6
      & 96.7 & 62.0 & \textbf{82.4} & 74.8 \\
    w/ PAWKL
      & \textbf{98.7} & \textbf{99.7} & 96.3 & \textbf{98.0} & \textbf{98.2}
      & \textbf{60.0} & \textbf{58.9} & 88.9 & \textbf{97.0}
      & \textbf{97.1} & \textbf{84.4} & 76.3 & \textbf{80.4} \\
    \bottomrule
  \end{tabular*}
\end{table*}

\begin{table}[!t]
  \centering
  \caption{Effect of frequency regularization and decay weighting on
  LIBERO success rates (\%). A decay coefficient of 0 retains
  $\mathcal{L}_{\mathcal{H}}$ with uniform frequency weights.}
  \label{tab:fr_decay_ablation}
  \small
  \setlength{\tabcolsep}{2pt}
  \begin{tabular*}{0.8\columnwidth}{@{\extracolsep{\fill}}lrrrrr@{}}
    \toprule
    Frequency loss setting & Spatial & Object & Goal & Long & Avg. \\
    \midrule
    w/o $\mathcal{L}_{\mathcal{H}}$
      & 98.0 & 99.2 & 95.8 & 93.0 & 96.5 \\
    \midrule
    Decay 0 (uniform)
      & 98.3 & 99.7 & \textbf{98.3} & 96.0 & 98.1 \\
    Decay 1 (exponential)
      & \textbf{98.7} & 99.7 & 96.3 & \textbf{98.0} & \textbf{98.2} \\
    Decay 2
      & 98.3 & \textbf{100.0} & 96.3 & 96.0 & 97.7 \\
    \bottomrule
  \end{tabular*}
\end{table}

\section{Conclusion}
We present \textbf{FreshVLA}, a frequency-guided, dynamics-sensitive framework for learning temporally hierarchical continuous latent actions for VLA policies. FreshVLA uses a high-pass prior to emphasize manipulation-relevant visual changes, while cross-timescale construction and frequency regularization encourage complementary representations of task progress and execution dynamics. For downstream control, a joint flow-matching policy denoises latent action and robot action streams using targets from a frozen latent action encoder. With 2B parameters, FreshVLA achieves 98.2\% average success on LIBERO, 98.0\% on LIBERO-Long, and an average sequence length of 4.5 on CALVIN. On zero-shot LIBERO-PLUS, it attains 80.4\% overall success and outperforms GR00TN1.7. Representation analyses show that FreshVLA’s latent actions are more robust to visual perturbations and yield lower regression error on long action sequences. Across four real-world manipulation tasks, FreshVLA outperforms $\pi_{0.5}$ by 7.5 percentage points in mean success rate. Together, these findings support its effectiveness for robust, long-horizon manipulation.

\bibliographystyle{unsrtnat}
\bibliography{main}

\newpage
\appendix

\section*{Supplementary Material}

\section{Implementation Details}
\label{sec:appendix_details}

\subsection{Dataset Composition Details}
\label{subsec:app_dataset}
To pretrain temporally hierarchical latent actions, we construct a diverse corpus using the LeRobot framework. The resulting mixture comprises 227,894 episodes and roughly 27.87 million frames. The detailed statistics of our pretraining data mixture are provided in Table~\ref{tab:app_dataset_composition}.

For real-world interaction dynamics, we aggregate three prominent datasets from the Open X-Embodiment (OXE)~\citep{oneill2024openx} repository. Specifically, DROID~\citep{khazatsky2024droid} provides the largest share of visual dynamics with 74,740 episodes and 19.82 million frames. RT-1~\citep{brohan2023rt1} contributes an additional 86,397 episodes with 3.72 million frames, while Bridge~\citep{bridge} supplies 28,935 episodes containing 982 thousand frames. To further broaden the coverage of real-world manipulation, we incorporate data from RoboMIND2~\citep{hou2026robomind2} and RoboCOIN~\citep{robocoin}. We sample episodes across all their constituent sub-datasets. This approach yields 2,621 episodes with 203 thousand frames for RoboMIND2, and 15,638 episodes with 1.79 million frames for RoboCOIN.

For structured and temporally dense dynamics, we include two simulated benchmarks: CALVIN~\citep{mees2022calvin} with 17,870 tabletop episodes and 1.07 million frames, and LIBERO~\citep{liu2023libero} with 1,693 language-conditioned episodes and 273 thousand frames. Ultimately, this mixture exposes the model to heterogeneous temporal scales: from short, dense simulated transitions to long-tailed real-world trajectories, which is essential for learning both fine-grained execution and long-horizon compositional structures.

\begin{table}[ht]
\centering
\caption{Detailed overview of the pretraining dataset mixture. The corpus aggregates real-world and simulated manipulation data to provide heterogeneous temporal scales for latent action learning.}
\label{tab:app_dataset_composition}
\resizebox{\textwidth}{!}{%
\begin{tabular}{@{}lrrrrrrrr@{}}
\toprule
\textbf{Metric} & \textbf{DROID} & \textbf{RT-1} & \textbf{Bridge} & \textbf{RoboCOIN} & \textbf{RoboMIND2} & \textbf{CALVIN} & \textbf{LIBERO} & \textbf{Total} \\ \midrule
\textbf{Episodes} & 74,740 & 86,397 & 28,935 & 15,638 & 2,621 & 17,870 & 1,693 & 227,894 \\
\textbf{Frames} & 19,823,040 & 3,724,262 & 982,895 & 1,790,845 & 203,049 & 1,071,743 & 273,465 & 27,869,299 \\ \bottomrule
\end{tabular}%
}
\end{table}

\subsection{Latent Action Architecture Details}
\label{subsec:app_architecture}

\begin{table}[ht]
\centering
\caption{Detailed architectural hyperparameters for the latent action model.}
\label{tab:app_architecture_details}
\resizebox{0.85\textwidth}{!}{%
\begin{tabular}{@{}llc@{}}
\toprule
\textbf{Module} & \textbf{Hyperparameter} & \textbf{Value} \\ \midrule
\multirow{2}{*}{\textbf{Inputs \& Backbone}} & Observation window size \& Resolution & 36 frames, $224 \times 224$ \\
 & Visual backbone & DINOv3 ViT-B/16 (Frozen) \\ \midrule
\multirow{2}{*}{\textbf{DynaAdapter $A_\psi$}} & Output token dimension & 768 \\
 & Cross-attention mechanism & 4 heads, single shared query \\ \midrule
\multirow{3}{*}{\textbf{Hierarchical Encoder $\mathcal{H}$}} & Latent-action space dimension & 64 \\
 & Substage count ($K$) & 6 \\
 & Temporal convolution (Conv1D) & Kernel size 3 (symmetric padding) \\ \midrule
\multirow{2}{*}{\textbf{Bidirectional Predictor $P_\eta$}} & Architecture & Causal Transformer \\
 & Layers \& Hidden dimension & 3 layers, 1024 \\ \bottomrule
\end{tabular}%
}
\end{table}

\textbf{Network Architecture.} 
The core architectural hyperparameters for our latent action model are summarized in Table~\ref{tab:app_architecture_details}. Specifically, the DynaAdapter $A_\psi$ aggregates spatial patches into frame-level representations $\mathbf{h}_t$ via cross-attention, which is supervised by the PAWKL loss using high-pass filtered frequency components. The hierarchical continuous encoder $\mathcal{H}$ utilizes symmetric padding in its 1D temporal convolution to integrate bidirectional temporal context, while the substage-scale representation is aligned with the low-frequency DCT bins via $\mathcal{L}_{\mathcal{H}}$.

\textbf{Sketched Isotropic Gaussian Regularization (SIGReg).}
Let $\mathbf{h}\in\mathbb{R}^{B\times T\times D}$ denote the frame-level
representations. At each time step, SIGReg computes empirical
characteristic functions across the batch for randomly sampled,
unit-norm projection directions. For direction $\mathbf{a}_m$, define
\begin{equation}
\begin{aligned}
\widehat{\phi}_{t,m}(\tau_q)
&=
\frac{1}{B}
\sum_{b=1}^{B}
\exp\!\left(
i\tau_q\,\mathbf{a}_m^\top\mathbf{h}_{b,t}
\right),
\\
\mathcal{L}_{\mathrm{SIG}}
&=
\frac{1}{TM}
\sum_{t=1}^{T}
\sum_{m=1}^{M}
B
\sum_{q=0}^{Q-1}
\omega_q
\left|
\widehat{\phi}_{t,m}(\tau_q)
-
\exp\!\left(-\frac{\tau_q^2}{2}\right)
\right|^2 .
\end{aligned}
\label{eq:app_sigreg}
\end{equation}
Here, $\mathbf{a}_m\in\mathbb{S}^{D-1}$ is a sampled unit-norm direction,
$i$ is the imaginary unit, and
\begin{equation}
\begin{aligned}
\tau_q &= \frac{3q}{Q-1},
\qquad
\Delta\tau = \frac{3}{Q-1},
\\
\omega_q
&=
c_q\,\Delta\tau\,
\exp\!\left(-\frac{\tau_q^2}{2}\right),
\qquad
c_q =
\begin{cases}
1, & q=0 \text{ or } q=Q-1,\\
2, & \text{otherwise}.
\end{cases}
\end{aligned}
\end{equation}
The weights implement trapezoidal quadrature over $\tau_q\in[0,3]$,
using the even symmetry of the Epps--Pulley integrand. The implementation
uses $Q=17$ quadrature points and $M=1024$ projection directions.

\subsection{Training Pipelines and Optimization}
\label{subsec:app_training_details}

\textbf{Implementation Framework.} 
Our downstream policy is built upon the StarVLA framework~\citep{community2026starvla}, utilizing Qwen3.5-2B~\citep{qwen2026qwen35} as the visual-language backbone to extract multimodal context. All models are trained on a single computation node equipped with 8 NVIDIA RTX A100 GPUs and evaluated on an NVIDIA RTX 4090 GPU.

\textbf{Stage I: Latent Action Pretraining.} 
The upstream model processes 36-frame windows with a substage resolution of $K=6$. We optimize the network for 120,000 steps using AdamW~\citep{loshchilov2019adamw} with a total batch size of 128 and a base learning rate of $10^{-4}$.

\textbf{Stage II: Joint Flow-Matching Policy.} 
During downstream policy training, the pretrained latent action encoder remains frozen. The DiT-B action expert receives an 8-dimensional robot proprioceptive state and an interleaved sequence of continuous latent action and 7-dimensional robot-action tokens, $\bigl(z_i^{\mathrm{token}}, a_i^{\mathrm{token}}\bigr)_{i=1}^{12}$, with an action horizon of $H_{\mathrm a}=12$. This stage is optimized with AdamW at a base learning rate of $10^{-4}$.

\section{Additional Results}
\label{sec:appendix_exp}

\subsection{Attention Analysis} 
\begin{figure}[t]
\centering
\includegraphics[width=\linewidth]{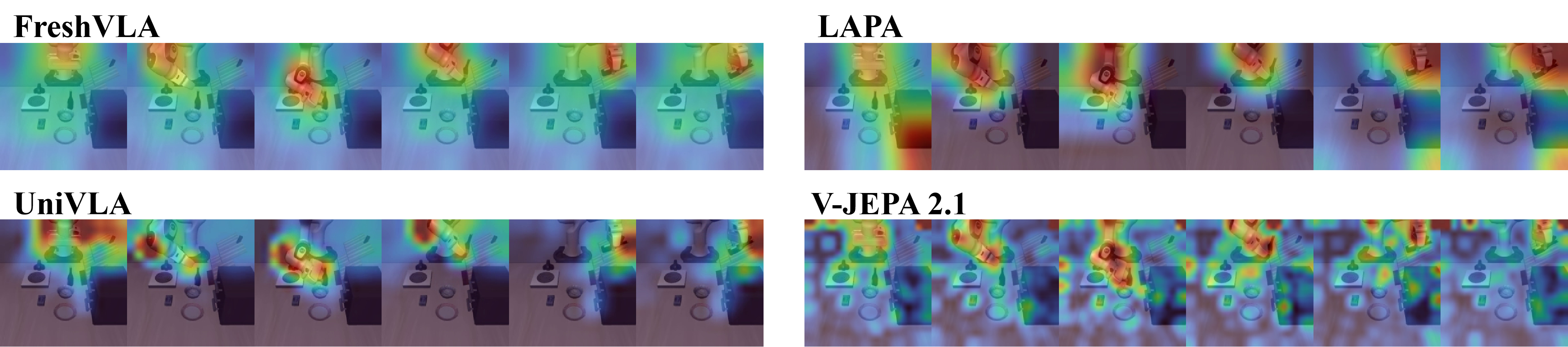}\\
\setlength{\abovecaptionskip}{2pt}
\caption{\ljm{Spatial attention maps across consecutive frames for FreshVLA, LAPA, UniVLA, and V-JEPA 2.1, highlighting differences in spatial focus and temporal consistency during manipulation.}}
\label{fig:attention-map}
\end{figure}

We visualize the visual attention patterns of V-JEPA 2.1~\citep{murlabadia2026vjepa21}, LAPA~\citep{ye2025lapa}, UniVLA~\citep{bu2025univla}, \ljm{ and FreshVLA by extracting model-specific attention weights over spatial
visual tokens and mapping them onto frame-wise heatmaps.} This allows us to compare how the models attend to task-relevant objects and interaction regions. As shown in Figure~\ref{fig:attention-map}, V-JEPA 2.1 attends broadly to the scene, including background regions and objects unrelated to the interaction. LAPA and UniVLA focus more on end-effector–object interactions, but their attention shifts noticeably across consecutive frames. In contrast, FreshVLA more consistently tracks the interacting object as it moves. This evidence suggests a more temporally coherent, interaction-aware representation that may better support long-horizon manipulation.

\subsection{Representation and Policy Interface}
To isolate the contribution of the latent action representation to downstream control, we first vary the latent action model while holding the joint denoising interface fixed. As shown in Table~\ref{tab:rep_libero}, replacing VILLA-X with FreshVLA increases mean LIBERO success from 94.8\% to 98.2\%, including an 11.8-point gain on LIBERO-Long. Success also increases in each LIBERO-PLUS category, with the mean rising from 74.3\% to 80.4\%. These results show that changing the latent action representation alone can improve both long-horizon control and robustness to perturbations. We then hold the FreshVLA representation fixed and vary the policy interface. Following the published post-training principle of JALA~\citep{luo2026jala}, our JALA-style implementation aligns VLM hidden states to latent-action targets and conditions robot-action denoising on those features. Feature alignment achieves the highest standard LIBERO mean (98.4\%). With the latent action held fixed, joint denoising outperforms JALA-style conditioning on LIBERO-Long (98.0\% vs. 96.0\%) and achieves higher macro-average success on LIBERO-PLUS (80.4\% vs. 79.3\%), with gains of 5.8 and 4.7 percentage points under camera viewpoint changes and sensor noise, respectively. Both interfaces improve on the action-only baseline, suggesting that FreshVLA’s latent actions benefit downstream control under either policy interface.

\begin{table*}[t]
\centering
\captionsetup{font+={color=blue}}
\caption{Success rates (\%) on LIBERO and zero-shot LIBERO-PLUS
across latent representations and policy interfaces.
Robot denotes changes in the robot's initial state.
Bold indicates the best result in each column before rounding.}
\label{tab:rep_libero}
\label{tab:rep_plus}

\small
\setlength{\tabcolsep}{2pt}
\renewcommand{\arraystretch}{1.08}
\begin{tabular*}{\textwidth}{
  @{\extracolsep{\fill}}l*{5}{r}@{\hspace{1em}}*{8}{r}@{}
}
\toprule
& \multicolumn{5}{c}{\textbf{LIBERO}}
& \multicolumn{8}{c}{\textbf{LIBERO-PLUS (zero-shot)}} \\
\cmidrule(lr){2-6} \cmidrule(lr){7-14}
Method
  & Spa. & Obj. & Goal & Long & Avg.
  & Cam. & Robot & Lang. & Light & Bg. & Noise & Layout & Avg. \\
\midrule
QwenGR00T
  & 98.7 & 96.0 & 96.3 & 93.0 & 96.0
  & 58.9 & 58.8 & 86.0 & 91.8 & 84.0 & 71.8 & 76.1 & 75.3 \\
FreshVLA, JALA
  & \textbf{99.7} & \textbf{100.0} & \textbf{98.0} & 96.0 & \textbf{98.4}
  & 54.2 & \textbf{63.8} & \textbf{89.1} & 95.8 & 96.0 & 79.7 & 76.2 & 79.3 \\
VILLA-X, JFM
  & 97.8 & 99.4 & 95.6 & 86.2 & 94.8
  & 45.9 & 55.9 & 83.9 & 94.1 & 93.2 & 71.0 & 76.3 & 74.3 \\
FreshVLA, JFM
  & 98.7 & 99.7 & 96.3 & \textbf{98.0} & 98.2
  & \textbf{60.0} & 58.9 & 88.9 & \textbf{97.0}
  & \textbf{97.1} & \textbf{84.4} & \textbf{76.3} & \textbf{80.4} \\
\bottomrule
\end{tabular*}
\end{table*}

\subsection{Policy Inference Cost}
We compare robot-action-only and joint denoising using the same Qwen3.5-2B backbone and DiT-B action head on 500 dual-view LIBERO observations spanning 40 tasks. Both configurations run on a single NVIDIA H200 GPU with batch size 1, a 12-step action horizon, an 8-dimensional state, and four denoising steps. Timing covers preprocessing, VLM encoding, denoising, and output transfer. Disk access, RPC, action unnormalization, and robot execution are excluded. Joint denoising increases median latency by 1.58 ms (1.85\%) and P95 latency by 1.39 ms. The additional 0.756 MiB allocated at the action-head peak remains below the VLM-stage peak, leaving the measured overall peak unchanged (Table~\ref{tab:inference_cost}). Under this evaluation protocol, jointly predicting latent and robot actions increased both median and 95th-percentile policy-call latency by less than 2 ms, suggesting modest inference overhead for real-world deployment on comparable hardware.
\par\medskip
\begingroup
\captionsetup{font+={color=blue}}
\centering\small
\captionof{table}{Model-side inference cost per policy call on one H200 GPU.}
\label{tab:inference_cost}
\setlength{\tabcolsep}{9pt}
\renewcommand{\arraystretch}{1.08}
\begin{tabular}{@{}lrr@{}}
\toprule
Metric & \shortstack[r]{Robot actions\\only} & \shortstack[r]{Joint latent/\\robot actions} \\
\midrule
Median latency (ms) & 85.22 & 86.80 \\
P95 latency (ms) & 87.63 & 89.02 \\
Mean latency (ms) & 86.58 & 87.23 \\
Peak GPU allocation (GiB) & 4.878 & 4.878 \\
\bottomrule
\end{tabular}
\par
\endgroup

\section{Real-World Generalization Analysis}
\label{subsec:app_real_world_generalization}

\begin{figure}[!t]
  \centering
  \setlength{\abovecaptionskip}{2pt}
  \includegraphics[width=0.92\linewidth]{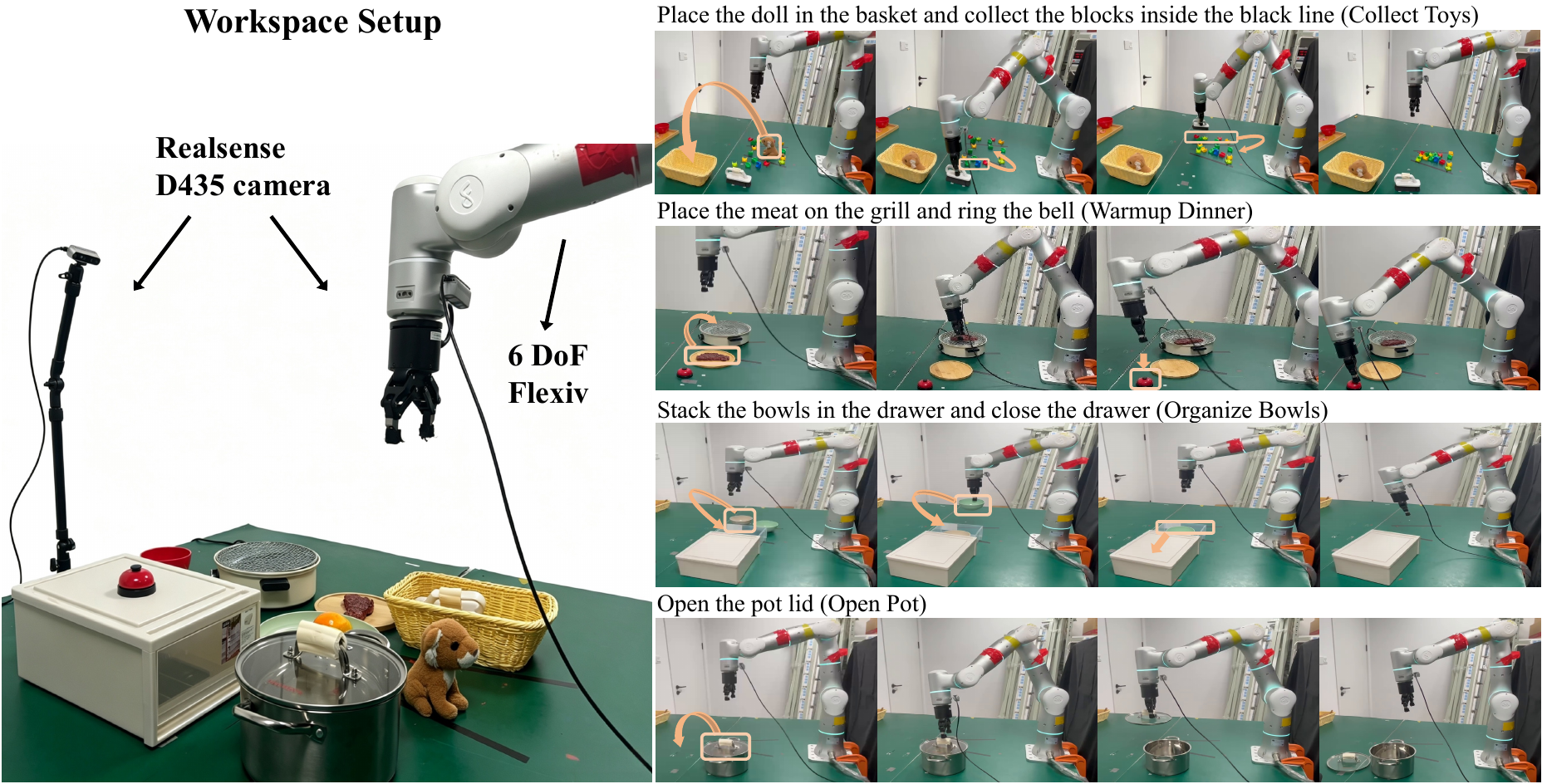}
  \caption{Real-world FLEXIV setup and representative manipulation sequences.
    Left: the 6-DoF Flexiv arm with a RealSense D435 camera.
    Right: representative rollouts for Collect Toys, Warmup Dinner,
    Organize Bowls, and Open Pot.}
  \label{fig:real_world_setup}
\end{figure}

\begin{figure}[!t]
  \centering
  \setlength{\abovecaptionskip}{2pt}
  \includegraphics[width=0.92\linewidth]{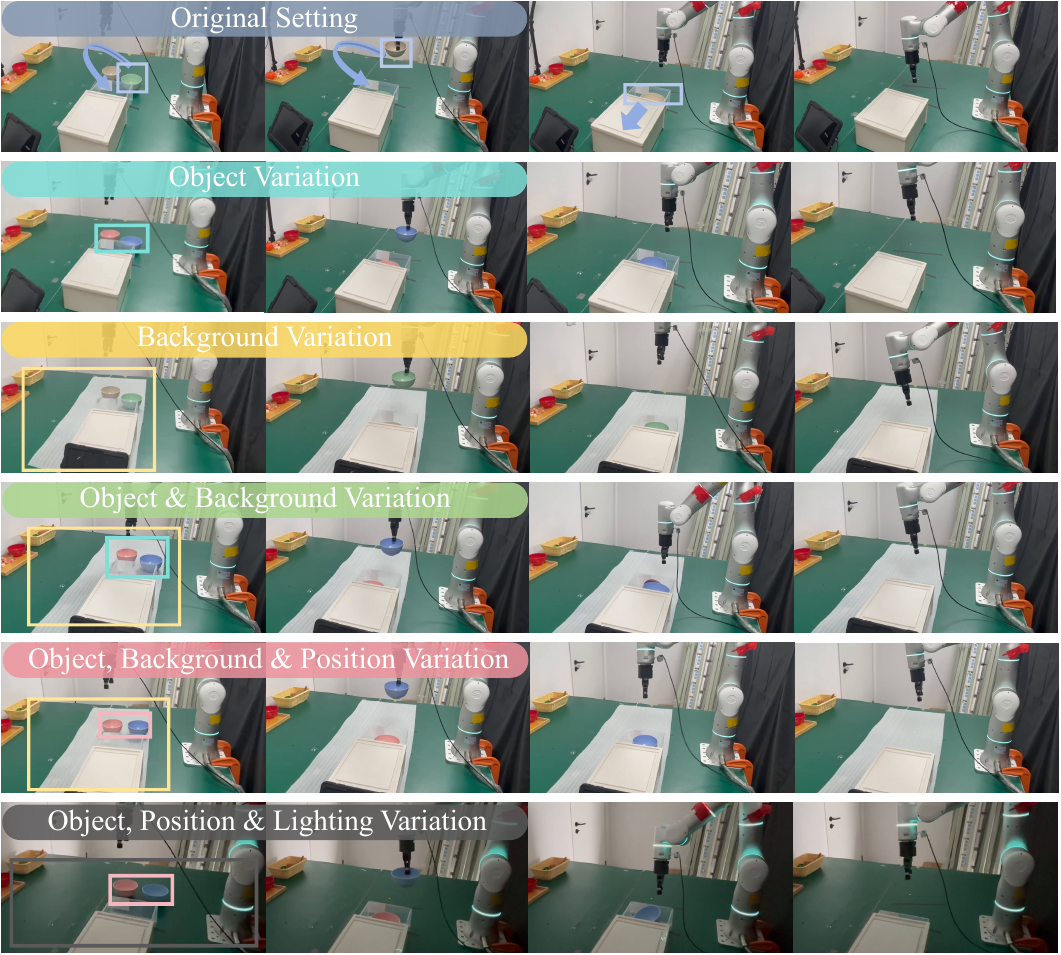}
  \caption{Real-world generalization evaluation on the Organize Bowls task.
    The sequences illustrate FreshVLA successfully completing the multi-stage
    manipulation under compounding out-of-distribution perturbations, including
    variations in object appearance, background texture, object position,
    and environmental lighting.}
  \label{fig:app_real_world_generalization}
\end{figure}

To further assess the generalization capability of FreshVLA in real-world scenarios, we conduct qualitative evaluations under various out-of-distribution visual and physical shifts. As depicted in Figure~\ref{fig:app_real_world_generalization}, we introduce multiple compounding perturbations to the original tabletop setting for the \textbf{Organize Bowls} task. These testing conditions include:
\begin{itemize}
    \item \textbf{Object Variation}: Altering the color, shape, and semantic identity of the manipulated objects.
    \item \textbf{Background Variation}: Introducing a large white mat to change the background texture and layout compared to the original green tabletop.
    \item \textbf{Position Variation}: Shifting the initial locations of the target objects and the receptacle.
    \item \textbf{Lighting Variation}: Reducing the environmental illumination.
\end{itemize}

These qualitative observations are consistent with our quantitative results on the simulated LIBERO-PLUS benchmark. The rollouts show that FreshVLA completes the Organize Bowls task despite substantial visual distractors, even when multiple visual perturbations are combined, suggesting robustness to real-world visual variation.

\clearpage

\end{document}